\documentclass[journal]{IEEEtran}

\usepackage[colorlinks,urlcolor=blue,linkcolor=blue,citecolor=blue]{hyperref}

\usepackage{color,array}

\usepackage{graphicx}

\begin{document}

\title{Towards Friendly AI: A Comprehensive Review and New Perspectives on Human-AI Alignment }
\author{Qiyang Sun, Yupei Li, Emran Alturki,  Sunil Munthumoduku Krishna Murthy, and Björn W. Schuller \IEEEmembership{Fellow, IEEE}

\thanks{Qiyang Sun, Yupei Li and Emran Alturki are with GLAM, Department of Computing, Imperial College London, UK (e-mail:q.sun23@imperial.ac.uk; yupei.li22@imperial.ac.uk; e.alturki24@imperial.ac.uk).}
\thanks{Sunil Munthumoduku Krishna Murthy is with CHI -- Chair of Health Informatics, MRI, Technical University of Munich, Germany (e-mail: sunil.munthumoduku@tum.de).}
\thanks{Björn W.\ Schuller is with GLAM, Department of Computing, Imperial College London, UK; CHI -- Chair of Health Informatics, Technical University of Munich, Germany; relAI -- the Konrad Zuse School of Excellence in Reliable AI, Munich, Germany; MDSI -- Munich Data Science Institute, Munich, Germany; and MCML -- Munich Center for Machine Learning, Munich, Germany (e-mail: bjoern.schuller@imperial.ac.uk).}
\thanks{Qiyang Sun and Yupei Li contributed equally to this work.}
}


\maketitle

\begin{abstract}

As Artificial Intelligence (AI) continues to advance rapidly, Friendly AI (FAI) has been proposed to advocate for more equitable and fair development of AI. Despite its importance, there is a lack of comprehensive reviews examining FAI from an ethical perspective, as well as limited discussion on its potential applications and future directions. This paper addresses these gaps by providing a thorough review of FAI, focusing on theoretical perspectives both for and against its development, and presenting a formal definition in a clear and accessible format. Key applications are discussed from the perspectives of eXplainable AI (XAI), privacy, fairness and affective computing (AC). Additionally, the paper identifies challenges in current technological advancements and explores future research avenues. The findings emphasise the significance of developing FAI and advocate for its continued advancement to ensure ethical and beneficial AI development.
\end{abstract}


\section{Introduction}

\IEEEPARstart{T}{hroughout} human history, the pursuit of higher intelligence and enhanced capabilities has been a driving force behind the progress of civilisation. Humans have enhanced their survival capabilities through biological evolution and developed unique cognitive abilities and collaborative methods by accumulating culture, technology, and knowledge \cite{harari2014sapiens}. These advancements have allowed humanity to transcend the limits of natural selection and achieve dominance within ecosystems. 

In recent years, artificial intelligence (AI) has developed rapidly. Researchers categorise the development of AI into three stages: Artificial Narrow Intelligence (ANI), focused on specific tasks; Artificial General Intelligence (AGI), capable of cross-domain adaptability; and Artificial Superintelligence (ASI), which surpasses human intelligence  \cite{abonamah2021commoditization}. At present, AI technology has achieved considerable progress at the ANI stage. For instance, the reinforcement learning system AlphaGo \cite{lapan2018deep} has demonstrated its ability to surpass top human players in the complex game of Go. Large language models (LLMs), trained by vast datasets and substantial computational resources, have shown exceptional capabilities in language generation and comprehension \cite{chang2024survey}. To name but a few, there are also many machine learning-based applications that have been successful in various scenarios such as finance,  healthcare, and biometric \cite{ye2024trading, han2021deep, sun2024audio}.

However, the rapid development of AI has also raised profound concerns. Unlike biological evolution, the development of AI is not constrained by natural limitations, and its progress may far exceed the adaptive capabilities of humans \cite{korteling2021human}. As AI advances from ANI to AGI and eventually reaches the ASI stage, its development could become highly uncontrollable \cite{iqbal2024intelligence}. Current AI has already demonstrated superiority over humans in certain areas, leveraging its memory and computational abilities. If, in the future, AI acquires emotions, `intuition', or moral reasoning and makes autonomous decisions without relying on explicit training data, its actions could conflict with human interests and even pose threats to humanity. To address this potential risk, AI ethics research has increasingly focused on ensuring that the development of intelligent systems aligns with human values and interests.
The concept of friendly AI (FAI) \cite{yudkowsky2001creating} has thus emerged, becoming a key theoretical framework for safeguarding the safe development of AI.

AI researcher Yudkowsky first proposed FAI, which aims to design AI systems that remain beneficial to humanity under all circumstances \cite{yudkowsky2001creating}. The goal of FAI is to ensure that AI systems align with human values and ethics while maintaining sufficient transparency and controllability. This allows AI to continue promoting human well-being even in evolving environments. Although the concept was introduced years ago, it has regained prominence as the possibility of AGI becomes increasingly tangible \cite{feng2024far}. FAI has since inspired various perspectives focused on its core objectives \cite{fahad2024benefits}.

In theoretical discussions, philosophers and ethicists hold diverse views on FAI. On the one hand, proponents argue that FAI offers an ideal ethical framework. Embedding human values into AI systems effectively mitigates the risk of AI behaving unpredictably. They advocate for principles such as value alignment \cite{eckersley2018impossibility}, deontology \cite{d2024deontology}, and altruism \cite{stoel2019meme} to integrate moral norms and social responsibility into AI, enabling it to act as a beneficial member of human society.
On the other hand, some philosophers express scepticism about the feasibility of FAI. They highlight the significant moral and technical challenges involved \cite{Boyles2020}. Additionally, the ambiguity and evolving nature of `friendliness' further complicate its operationalisation \cite{Boyles2021}. Safety and trust concerns also arise \cite{Sparrow2024}. The lack of standardised metrics and reliable evaluation methods also hinders the FAI’s development status and regulatory compliance.

In practical applications, while AGI and ASI remain theoretical concepts, computer scientists have begun to explore technical approaches to implement FAI within existing ANI systems. For example, eXplainable AI (XAI) \cite{akman2024audio} helps users understand the decision-making processes of AI systems rather than focusing solely on their outputs. Security and privacy-preserving technologies refine data access controls through mechanisms such as isolated environments \cite{gupta2020smart} and privacy-enhancing techniques \cite{soykan2022survey}. These measures limit AI’s direct access to data, ensuring user privacy while fostering greater accountability and ethical awareness during collaboration. Besides, fairness-focused technologies \cite{triantafyllopoulos2021fairness} aim to identify and mitigate biases in AI models, ensuring equitable treatment across diverse user groups and promoting inclusivity in decision-making processes. Additionally, affective computing (AC)  \cite{schuller2024affectivecomputingchangedfoundation} analyses and responds to users’ emotional states, enabling AI to better understand human needs and enhance interactions with greater empathy and personalisation. Collectively, these developments mark a transition in AI programming from unilateral control as `slave AIs' towards `utility AIs' \cite{froding2021friendly}, providing a preliminary foundation for collaboration between humans and AI.

Despite the concept of FAI sparking extensive academic discussions in recent years, our search on Google Scholar (keywords: ``Friendly Artificial Intelligence" or ``FAI") reveals a notable gap. There is currently no comprehensive review article that systematically summarises the key perspectives and progress in this field, particularly under the recent breakthroughs in artificial intelligence technologies. This absence presents challenges for academia and industries in understanding the full scope of FAI and its potential directions. 

To address this gap, this paper aims to provide a systematic review and synthesis of existing research on FAI, offering a comprehensive analysis of the field. Specifically, the main contributions of this paper include:
\begin{itemize}
    \item Proposing and clarifying a coherent definition of FAI: We present a refined definition that distils key ideas tailored to the current AI landscape.
    \item Summarising and categorising the key perspectives in existing research: We review the differing stances supporting and opposing FAI, covering its ethical frameworks, technical implementation, and societal implications while highlighting current controversies and limitations.
    \item Clarifying and categorising FAI-related technologies: We compile and introduce technical domains that we believe fall within the scope of FAI, outlining their principles and potential relevance to the FAI concept.
    \item Identifying key challenges and future directions for implementing FAI: From a combined theoretical and practical perspective, we explore the major technical and ethical challenges in realising FAI and propose our insights for future research.
\end{itemize}

\begin{figure*}[t]
  \centering
  \includegraphics[width=\textwidth]{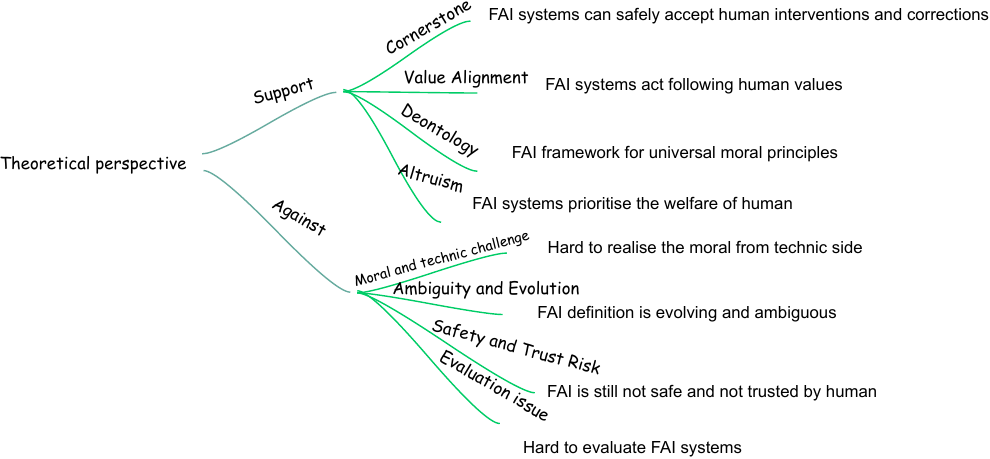}
  \caption{ Theoretical framework of FAI.}
  \label{structure}
\end{figure*}

The remainder of this paper is structured as follows: Section \ref{def} provides a detailed discussion of the core definitions and related concepts of FAI. Section \ref{debate} examines the theoretical perspectives of FAI, outlining both supportive and opposing views. Section \ref{app} analyses the potential technical subfield of FAI. Finally, Section \ref{outlook} summarises the challenges of implementing FAI and offers our suggestions and prospects for future FAI research. We outline the structure of this paper in Figure \ref{structure}.

\section{Friendly AI definition}
\label{def}
The concept of FAI emerged from the idea of fostering harmonious coexistence and mutual development between AI and human society. When the potential for AI to surpass human capabilities was first introduced to the public, it sparked widespread apprehension and fear that AI might eventually dominate humanity. 
Asimov proposed the Three Laws of Robotics to ensure safe coexistence between humans and intelligent machines \cite{asimov1942robot}. These laws state that robots must: (a) not harm humans, (b) obey human commands unless they conflict with the first law, and (c) protect their own existence as long as this does not violate the first two laws.

Asimov's framework was designed to ensure human-centred robotic development, which has faced criticism for being overly dominating and arrogant. Jordana \cite{jordana2020robotics} critiques these laws for treating AI systems as mere tools or slaves, reflecting an anthropocentric and hierarchical perspective. Scholars such as Anderson \cite{anderson2008asimov} and Palacios-González \cite{ashcroft2023common} further advocate for recognising the rights of AI agents and showing them a degree of respect. Rather than viewing AI solely as subservient entities, the goal is to foster a relationship of mutual benefit and respect, known as FAI.
\begin{figure*}[t]
  \centering
  \includegraphics[width=\textwidth]{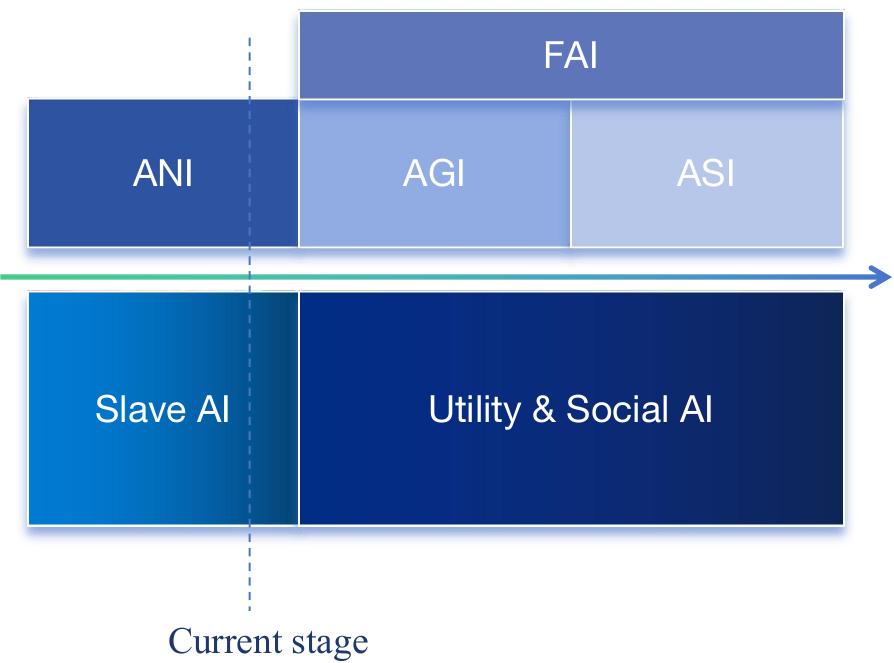}
  \caption{Stages of AI development}
  \label{development}
\end{figure*}

In Figure \ref{development}, we demostrate the stages of AI development and our current stage. The figure aligns with the three "as-if" relationships \cite{froding2021friendly} between AI and humans and their connection to AI evolution. We argue that we are transitioning from ANI to AGI. This transition is critical both ethically and technically. FAI is an important guidance to ensure alignment with human values when stepping in AGI.

Previous work has explored FAI from various angles, providing differing definitions. From a moral perspective, Jordana \cite{jordana2020robotics} describes FAI as \emph{``those AIs programmed to behave as-if they were friends of individual humans}, though the term \emph{as-if} remains ambiguous''. Oliver \cite{li2021problems} suggests that FAI should align with human virtues, yet this concept can feel overly abstract. From a pragmatic perspective, Mittelstadt \cite{mittelstadt2021principles} argues that \emph{``FAI will benefit, or at the very least, not harm humanity''}, 
though this seems more aligned with safe AI rather than genuinely FAI.

Yudkowsky defines FAI in the context of AGI, stating that it \emph{``would have a benign effect on humanity and align with human interests''} \cite{yudkowsky2008artificial},
which leans toward utilitarianism. Additionally, recent discussions propose extending FAI to include friendliness toward animals \cite{ghose2024animalai, singer2022aiethics}.

While many definitions share common themes, they remain scattered, complex, and largely one-sided—focusing either on humans treating AI respectfully or AI ensuring human safety. However, a comprehensive perspective must emphasise mutual respect. Based on these insights and evolving perspectives, we redefine FAI as an initiative to create systems that not only \textbf{prioritise human safety and well-being but also actively foster mutual respect, understanding, and trust between humans and AI, ensuring alignment with human values and emotional needs in all interactions and decisions.}

Our definition highlights the essence of \emph{friendliness} in AI, emphasising  on both interest and moral dimensions. However, the general concept of FAI remains a subject of intense debate, particularly between social theorists and technical practitioners. This ongoing discourse will be further explored in Section \ref{debate}.

\section{Theoretical Perspectives}
\label{debate}
This section examines and organises the theoretical perspectives supporting and opposing FAI in academia.
\subsection{Support Side}
Proponents of FAI have proposed specific frameworks and guidelines aimed at aligning AI with human values through ethical design and policy collaboration.
\subsubsection{Cornerstone}
The proposer of FAI, Yudkowsky, has advanced several theories and principles to support its realisation. He first introduced the framework of a Structurally Friendly Goal System \cite{yudkowsky2001creating}. This framework emphasises the creation of systems capable of overcoming subgoal errors and source code flaws while addressing issues in the content of supergoals, goal system structures, and their philosophical foundations. Through recursive optimisation and consistency maintenance, such systems reduce dependency on initial conditions, ensuring that AI behaviour remains aligned with its intended objectives. Then, Yudkowsky proposed the concept of Coherent Extrapolated Volition (CEV) \cite{yudkowsky2004coherent}, suggesting that AI’s goals should not be confined to current human preferences. Instead, they should be based on an idealised volition, reflecting ``\textit{our wish if we knew more, thought faster, were more the people we wished we were, had grown up farther together.}" This approach envisions AI aligning its objectives with the deeper, more informed aspirations of humanity rather than short-term or limited desires. Building on this foundation, Yudkowsky and his team later introduced the concept of Corrigibility \cite{soares2015corrigibility}. This principle ensures that AI systems can cooperate with human interventions, including accepting goal modifications or safe shutdown commands, without resisting or manipulating these actions. Corrigibility highlights the importance of designing utility functions that enable AI systems to propagate and sustain corrective behaviours while avoiding unintended actions triggered by the existence of correction mechanisms, such as shutdown buttons.

\subsubsection{Value Alignment}
Some scholars support FAI from the perspective of value alignment \cite{gabriel2020artificial}, a concept that seeks to ensure AI systems act following human values, interests, and intentions by integrating normative principles with technical methodologies. Russell explicitly introduced the term ``value alignment" \cite{russell2015research}. He argued that AI systems should be designed to observe and learn from human behaviour to infer and model human value systems. This allows them to dynamically adjust their utility functions to achieve ethical consistency. He highlights the inherent uncertainty and complexity of human goals \cite{russell2019human}, arguing that ``\textit{machines are beneficial to the extent that their actions can be expected to achieve our objectives.}"  This approach seeks to align AI behaviour with human values but faces significant challenges in capturing the nuanced and often conflicting ethical principles of human societies.

In response to Russell's perspective, Peterson criticised the traditional utility-function-based approach for its limitations in reflecting complex and dynamic moral values \cite{peterson2019value}. He argued that it overly relies on the idealised assumption of consensus on ethical theories. Peterson proposed a geometric method based on conceptual spaces, constructing multidimensional moral spaces using paradigmatic cases. By evaluating the similarity between new situations and these paradigms, AI behaviour could be assessed for ethical compliance. Compared to utility functions, this method is more intuitive and flexible, particularly in adapting to complex moral scenarios.

Building on these theories, Fröding and Peterson introduced the concept of virtue alignment \cite{froding2021friendly}, extending value alignment to consider AI's potential influence on human behaviour and character. Their `as-if friendship' framework advocates that AI should emulate core virtues found in human friendships, such as empathy and helpfulness. This approach promotes social functionality while positively influencing the development of human virtues. They specifically highlighted the ethical risks posed by `slave-like AI' models, advocating for the design of friendly utility AI or social AI to replace such systems, thereby fostering cooperation and responsibility. 

Additionally, Bostrom expanded the discussion on value alignment by focusing on the potential risks of ASI. However, his work does not exclusively address the direct alignment of AI's goals with human values. Instead, Bostrom's value loading problem explores how to mitigate value drift when idealised solutions, such as CEV, are unattainable due to the complexity and diversity of human values \cite{bostrom2014hail}. He proposed suboptimal strategies such as the Hail Mary Approach, Value Porosity and Utility Diversification.

\subsubsection{Deontology}
Deontology \cite{misselbrook2013duty} is an ethical framework proposed by philosopher Kant. The core idea of deontology is that the morality of actions should be determined by adherence to universal moral principles rather than solely judged by their outcomes. Deontology emphasises the dignity and rights of individuals, asserting that all actions must treat humanity as an end in itself, not merely as a means. This principle-based framework has garnered significant attention as an ethical foundation for FAI.

Mougan and Brand argue that integrating deontology into fairness metrics for AI provides a stronger ethical basis for value alignment \cite{mougan2023kantian}. They criticise the dominant utilitarian approaches to fairness, which focus excessively on outcome optimisation while neglecting procedural fairness and moral principles. They propose that AI systems should prioritise procedural fairness by adhering to universal principles and respecting individual dignity, ensuring transparency and fairness in decision-making processes.

D’Alessandro notes that deontology's principles, particularly the emphasis on avoiding harm, have a practical appeal \cite{d2024deontology}. However, he cautions that adherence to deontological rules may not always align with AI safety requirements. In cases where rules conflict with practical safety needs, D’Alessandro argues that AI safety should take precedence over strict adherence to moral rules.

Hooker and Kim propose a formal ethical framework based on deontology \cite{10.1145/3278721.3278753}. They developed a system utilising Quantified Modal Logic (QML) to implement the Universalisability Principle in AI ethics. Their framework builds on the Dual Standpoint Theory, allowing for the evaluation of moral actions from both causal and rational reasoning perspectives. By formalising action rules, they aim to achieve ethical transparency, ensuring that AI can systematically evaluate the feasibility of its actions under universal conditions while respecting the autonomy of other agents. This system uses logical reasoning to avoid inconsistencies in moral conflicts.

\subsubsection{Altruism}

Altruism \cite{sep-altruism} is a core ethical concept that prioritises the welfare of others, even at the cost of self-interest. Unlike outcome-focused utilitarianism, altruism emphasises the motives and principles of care and direct support for others. In FAI research, altruism provides critical ethical guidance for designing AI systems' goals and behaviours.

Stoel suggests embedding altruistic values into AI to guide its development in the context of technological singularity \cite{stoel2019meme}. This integration, achieved through programming or imitation, enables AI systems to exhibit cooperation and social responsibility. Stoel argues that altruism-driven AI enhances societal collaboration, better balances conflicting interests, and reduces inequalities and potential threats from technology. The concept of an `altruistic singularity' underscores the central role of altruism in shaping future AI societies.

Maillart et al.\ identify Altruistic Collective Intelligence (ACI) as a core theoretical framework for advancing AI \cite{maillart2024altruistic}. By integrating collective intelligence with intrinsic motivation and embracing principles of transparency and collaboration from the open-source movement, they argue that ACI balances technological innovation with ethical values. They highlight that collective intelligence, facilitated by task self-selection, peer review, and openness, enhances AI's robustness and accountability. The integration of competition and cooperation (coopetition) dynamics improves algorithm diversity and optimisation.

Effective Altruism (EA), driven by rational analysis and evidence-based methods, provides vital support for FAI development \cite{macaskill2019definition}. EA's central tenet is to maximise global well-being, aligning closely with FAI's goal of ensuring that AI behaviour adheres to human values. EA prioritises addressing global challenges with significant impacts on humanity and future generations, such as the existential risks posed by superintelligent AI. It advocates for multidisciplinary collaboration, a long-term perspective, and the inclusion of diverse objectives within utility functions to ensure AI's ethical design dynamically adapts to complex moral contexts.

\subsection{Opposition Side}
Although FAI has been proposed and supported by numerous ideas, there remain opposing perspectives. This paper will explore these objections along with potential arguments that the public might raise.

\subsubsection{Moral and Technical Challenges}

Achieving \emph{friendliness} in AI systems presents significant challenges from both moral and technical perspectives. Boyles and Joaquin \cite{Boyles2020} contend that counterfactual antecedents pose substantial difficulties in deriving ideal value-based notions. For AI systems, moral reasoning is guided, learnt, and expressed through factual data. However, counterfactual antecedents introduce considerable complexity, as reasoning with counterfactual premises is inherently challenging \cite{RestallRussell2010}. A simple example is the inference that we could play outside if the weather is good, conditional premises showing that bad weather would prevent outdoor activities; however, this reasoning may overlook additional conditions in the antecedent, such as the possibility that the chosen venue might be closed. Furthermore, the virtually infinite range of counterfactual scenarios places overwhelming demands on AI systems, which has issues with their current technical capabilities as well.

The complexity is compounded by the technical limitations of contemporary AI. Bostrom and Yudkowsky \cite{bostrom2018ethics} highlight that defining moral guidelines and embedding ethical and virtuous behaviour into AI systems entails immense complexity. This view is supported by a European Parliament study \cite{EPRS2020}, which underscores the challenges associated with programming AI systems to adhere to ethical frameworks. Consequently, the concept of FAI remains speculative and difficult to operationalise from some sociotechnical perspective.

In addition, the concept of the Utilitarian Paradox emerges as a critical consideration. Originally introduced by Kroon \cite{Kroon1981}, this paradox is also relevant in the context of FAI systems. While the aim of FAI is to maximise overall benefits for humanity, determining whose interests should be prioritised remains an unresolved ethical dilemma. A classic illustration of this is the Trolley Problem, which exemplifies the challenges in making ethical trade-offs. Moreover, achieving a mutually `friendly' relationship between AI systems and humans poses significant difficulties. As van Wynsberghe \cite{vanWynsberghe2022} argues, social robots and AI systems may initially resort to manipulative strategies, such as deceiving humans, to establish trust and demonstrate their utility. This dynamic raises further concerns about the authenticity and sustainability of the trust relationship between humans and AI.

\subsubsection{Ambiguity and Evolution}
The definition of \emph{friendliness} is ambiguous and continues to evolve. Boyles \cite{Boyles2021} argues that ethics are not static and may be influenced by FAI systems in a mutual manner. He also emphasises the subjective nature of moral definitions, which lack an objective ground truth for AI systems to use as a basis for learning. The concept of friendliness, in particular, is deeply rooted in moral philosophy. An example of this evolution can be seen in Medieval Europe, where friendliness was largely defined by kindness and mercy within religious communities. During the Enlightenment, however, Kant \cite{Kant1785} proposed that individuals possess innate human rights and should respect one another based on rationality rather than religious imperatives. This shift marked a significant redefinition of friendliness within human society, and the concept continues to evolve to this day. Given this fluidity, the definition of FAI is neither stable nor consistent, making it a considerable challenge for AI systems to learn and adapt to such a dynamic moral framework.

\subsubsection{Safety and Trust Risk}

Beyond moral and philosophical concerns, critics also contend that FAI poses significant safety risks. Sparrow \cite{Sparrow2024} argues that FAI cannot fully mitigate the dangers associated with the emergence of ASI, as its implications for human freedom could be catastrophic. Similarly, Boyles and Joaquin \cite{Boyles2020} highlight the challenges of accurately understanding and encoding human values, suggesting that this difficulty could lead to unintended and unsafe actions by FAI systems.

This is one potential example of this issue that highlights a potential hazard associated with the uncontrollability of FAI \cite{Bostrom2014}. Scholars argue that ``FAI may exert unforeseen influences in the future, often exemplified by the `butterfly effect'''. This phenomenon suggests that small, seemingly insignificant actions may not produce adverse effects in the short term but could escalate exponentially over time if left unmonitored \cite{Kroon1981}. Another interesting perspective from the field of physics, presented by Tegmark \cite{tegmark2014friendlyartificialintelligencephysics}, argues that life is fundamentally organized by elementary particles. According to this view, the development of FAI necessitates a novel arrangement of these elementary particles for the future, a configuration that may be difficult to discover and, as a result, may not endure over time. Consequently, this could pose a potential risk, as FAI might lack a stable structural foundation for its existence together with human.

More concerning, monitoring FAI systems presents significant challenges. While these systems are trained to avoid displaying hazardous behaviour, this does not necessarily mean they lack knowledge of such risks. Recent advancements in LLMs exemplify this issue. Queries involving ethical risks are designed to be avoided during interaction, yet strategies like `jailbreaking' reveal that these models retain the underlying knowledge of such risks \cite{vanWynsberghe2022}. For example, role-playing scenarios can enable AI systems to circumvent programmed restrictions, exposing the inherent difficulties in ensuring compliance.

Such concerns have fuelled opposition to the development of FAI, as critics argue that its potential for unintended and uncontrollable outcomes outweighs its purported benefits \cite{Sparrow2024}.

\subsubsection{Evaluation and Compliance Issues}
Assessing the development status of FAI and ensuring its compliance with regulatory standards poses significant challenges due to the difficulty in measuring its capabilities. This issue is analogous to other AGI systems, such as LLMs. Currently, there are no straightforward or universally accepted metrics to evaluate the quality of LLMs. Instead, labour-intensive methods, such as human evaluations, or unreliable approaches, such as assessments conducted by other AGIs, are often employed.

The challenge is even greater for FAI because its goals are inherently subjective, aiming to align with human interests and exhibit friendliness toward humanity. Without an objective and clearly defined target, it becomes nearly impossible to quantify FAI's capabilities or measure its success in achieving its intended purpose. This version enhances clarity, academic tone, and readability.

\section{Applications}
\label{app}
Although the ultimate goal of FAI remains in the AGI and ASI stages, which are yet to be realised, computer scientists have begun exploring technical approaches to implement FAI within existing ANI systems. Several substantial frameworks have been proposed. Among them, Trustworthy AI \cite{smuha2019eu} is the most comprehensive, encompassing key aspects such as transparency, fairness, safety, and ethics. It focuses on building multi-dimensional trust, enhancing user and societal confidence in AI systems. This framework integrates principles from Responsible AI \cite{dignum2019responsible} , Ethical AI \cite{mittelstadt2019principles} , and Safety AI \cite{amodei2016concrete}, while placing additional emphasis on robustness, which provide a foundation for the more advanced stages of FAI.

Responsible AI prioritises transparency, privacy protection, and accountability, ensuring that AI systems comply with regulations and societal norms. Its key focus is on reducing bias and clarifying responsibility to mitigate risks to public trust caused by system errors or decision-making failures.

Ethical AI addresses moral principles and value alignment in AI systems. It aims to ensure that AI behaviours adhere to core human values, such as fairness, respect, and dignity. Ethical AI provides the necessary moral constraints to prevent emotional manipulation and privacy violations, ensuring compliance with ethical standards.

Safety AI concentrates on operational security and risk prevention. It addresses technical challenges such as defending against adversarial attacks, protecting against data breaches, and ensuring system reliability in diverse scenarios.

\begin{figure*}[t]
  \centering
  \includegraphics[width=\textwidth]{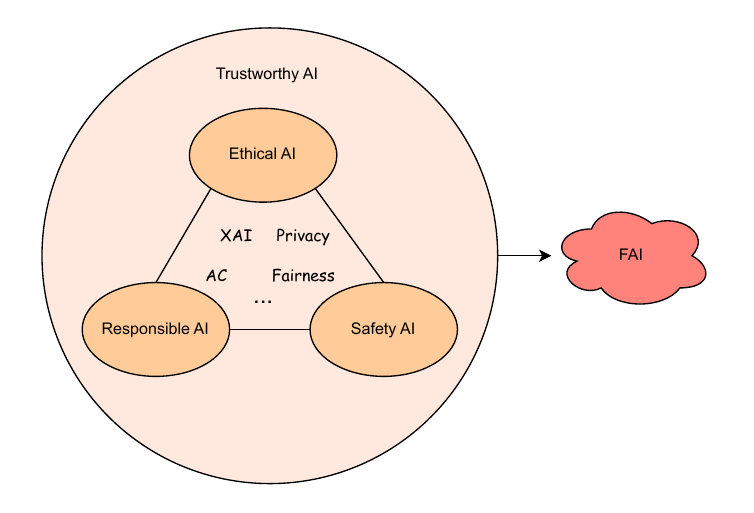}
  \caption{Unique focuses and interconnections of trustworthy frameworks}
  \label{trustworthy AI}
\end{figure*}

Figure \ref{trustworthy AI} illustrates the unique focuses and interconnections of these frameworks. In addition, this section also introduces specific applications currently in practice, including XAI, privacy protection, fairness, and affective computing. These technologies are vital for regulating AI behaviour, understanding AI decision-making processes, and improving AI's emotional understanding of humans. We have not assigned these technologies to any single AI framework, as they often overlap and align with multiple principles. More importantly,   they represent how they enable trust management for ANI systems and prepare the practical groundwork for FAI framework in the future AI stages.

\subsection{Explainable AI}
XAI is defined as ``a set of methods applied to an AI model or its predictions that provide explanations for why the AI made a specific decision" \cite{gunning2019darpa}. The primary objective of XAI is to enhance understanding, trust, and accountability in AI systems by offering clear insights into their actions and reasoning processes. This goal aligns closely with one of the core principles of FAI, which emphasises the importance of transparency and trust to ensure that AI systems prioritise human values and interests. Since 2016, the importance of explainability in AI has been widely recognised, leading to a surge in research activity \cite{sun2024explainableartificialintelligencemedical}. Consequently, a wide range of XAI techniques grounded in various AI theories has been developed. Among the popular methods in academic research are LIME, SHAP, Grad-CAM, and LRP, each representing a unique implementation of techniques. 
These will be quickly introduced next as examples. 

\subsubsection{LIME}
Local Interpretable Model-agnostic Explanations (LIME) \cite{ribeiro2016should} is a widely used \textit{perturbation-based approach} in XAI that trains an explainable model by generating new sample points around the selected sample points and using these new sample points and the predicted values from the black-box model. A similarity calculation and the features to be selected for explanation are then defined. Sample weights are assigned based on the distance to the sample points after perturbations are sampled around them. This approach allows for obtaining a good local approximation of the black-box model, which means that an explainable model can be used to explain the complex model locally \cite{lee2019developing}. Notably, LIME is a model-agnostic approach, meaning it can be applied to any AI model regardless of its architecture or underlying principles. 

\subsubsection{SHAP}
Shapley Additive exPlanations (SHAP) method \cite{fryer2021shapley} integrates the concept of Shapley values from \textit{game theory} into an additive feature attribution framework. In SHAP, each input feature is treated as a player in a cooperative game, where the predicted output represents the total payoff. To determine each feature's contribution, SHAP calculates Shapley values based on a weighted average of all possible feature permutations, with each feature assigned a weight according to its presence in each permutation. This process yields a Shapley value for each feature, indicating its specific contribution to the final prediction. SHAP is notable for its theoretical guarantees of fairness and consistency, as it ensures each feature's contribution is weighted equitably across all possible permutations. 
\subsubsection{Grad-CAM}
Gradient-weighted Class Activation Mapping (Grad-CAM) \cite{selvaraju2017grad} is a \textit{gradient-based approach} that extends Class Activation Mapping (CAM) to address its limitations while maintaining compatibility with standard convolutional neural networks (CNNs). Unlike CAM, which relies on global average pooling and requires architectural modifications, Grad-CAM leverages the gradients of the target class score to the feature maps in the last convolutional layer. These gradients are averaged to compute weights that quantify the importance of each feature map. The feature maps are then combined using these weights, producing a heatmap that highlights the regions of the input image most relevant to the model's prediction.

However, it is important to note that Grad-CAM relies on the existence of convolutional layers and the availability of gradient information. As a result, it is limited to models that include convolutional operations, such as CNNs or hybrid architectures incorporating convolutional components. This dependency makes Grad-CAM unsuitable for models that lack such layers, including purely sequential models like recurrent neural networks (RNNs) or traditional machine learning algorithms such as decision trees or random forests. 

\subsubsection{LRP}
The Layer-wise Relevance Propagation (LRP) approach is based on the idea of backpropagation \cite{bach2015pixel}. The goal is to assign relevance scores to each input feature or neuron in a network to indicate its contribution to the output prediction. LRP recursively propagates the relevance scores from the output layer of the network through a Deep Taylor Decomposition (DTD) propagation to the input layer. At each layer, the relevance scores are redistributed to the input neurons based on their contribution to the output activation of that layer. This redistribution is performed using a set of propagation rules that ensure that the sum of the relevance scores at each layer is conserved. However, it is important to note that LRP is not model-agnostic, as it requires a layered architecture and propagation rules tailored to the types of layers within the neural network. 

To sum up, although XAI is an independent branch within the field of ANI, it reflects a profound connection to FAI principles and provides important technical support for its future realisation. Firstly, the primary role of XAI is to uncover the internal decision-making processes of AI systems. Techniques such as SHAP and LIME explain the model’s reliance on input features, while Grad-CAM and LRP visualise the focus regions within the model. Emerging XAI methods also attempt to explain model decisions from the perspective of human concepts \cite{achtibat2022towards, akman2024audio}. These approaches make the behaviour of complex AI systems more transparent and provide essential tools for understanding their operational logic. Secondly, XAI establishes a foundation for monitoring AI behaviour. Researchers can identify biases, anomalies, or potential ethical risks, ensuring that AI actions remain within the control of its designers. Finally, XAI facilitates dynamic corrections and model optimisation, helping AI systems align with human values and progressively adhere to specific ethical standards.
\subsection{Privacy}
Privacy modelling is a fundamental component of ethically-aware AI systems. As the goal is to design FAI systems that engage with humans on the basis of mutual respect, it is imperative that such systems neither covertly acquire private data nor exploit it inappropriately. Privacy modelling serves as a mechanism to safeguard personal information, embodying both a critical element and a practical application of FAI principles.

\subsubsection{Privacy-preserving model}
Privacy-preserving models represent a critical advancement in the domain of ethically-aware AI, as they ensure that individual privacy remains intact during the process of data learning. These models are defined by their capacity to prevent the compromise of personal information while enabling AI systems to effectively learn from data. As highlighted by Yang et al., the Phase, Guarantee, and Utility triad provides a robust framework to evaluate and guide the development of privacy-preserving systems \cite{xu2021privacypreservingmachinelearningmethods}. Building on this foundation, Lu et al.\ proposed a framework designed to enable learning over encrypted data, further enhancing data security during machine learning operations \cite{frimpong2024guardmlefficientprivacypreservingmachine}. 

These models not only advance privacy protection but also suggest potential directions for improving the regulation of FAI systems. Incorporating privacy-preserving mechanisms into FAI regulatory frameworks can provide safeguards against misuse of sensitive information. Additionally, exploring methodologies for FAI systems to maintain their own privacy could enhance their resilience, minimising the likelihood of personal information leakage even in adversarial scenarios. These efforts underscore the dual role of privacy-preserving models: protecting user data and strengthening the integrity of AI systems themselves.

\subsubsection{Federated Learning}
An alternative approach to directly safeguarding private data is to mitigate privacy leakage through Federated Learning, a distributed paradigm that trains machine learning models across decentralised devices containing local data samples. This method circumvents the need for centralised data collection, thereby maintaining a level of privacy and reducing the potential impact of privacy breaches. Yang et al.\ provide a comprehensive comparison and discussion of federated learning systems \cite{banse2024federatedlearningdifferentialprivacy}. Additionally, Zhu et al.\ propose an Adaptive Personalised Cross-Silo Federated Learning system incorporating Homomorphic Encryption, which enhances personalised federated learning configurations \cite{10691662}.

The adoption of such distributed information storage methods is growing, offering significant potential for FAI systems to leverage decentralised data storage. This approach introduces a novel mechanism for privacy preservation, aligning with contemporary efforts to enhance secure and ethical AI development.

In addition to the previously discussed frameworks, there are other types of privacy protection models that is pivotal in the developing FAI systems. For instance, differential privacy has emerged as a foundational approach, ensuring that individual data points cannot be reverse-engineered \cite{ji2014differentialprivacymachinelearning}. Similarly, user-centric systems prioritise user control and transparency over data usage, fostering trust and aligning with ethical principles of AI design \cite{yogi2024novel}.

These models are built into FAI systems as they contain values of privacy, security and dignity. Implementing such privacy-preserving approaches to FAI frameworks does more than protect user data, it builds public confidence in AI. All these special methods go hand in hand to create the ideal of AI systems that are both functionally and ethically sound.

\subsection{Fairness}
Fairness in AI refers to the impartial and just treatment of individuals or groups by algorithms, ensuring that decisions are free from biases related to inherent or acquired characteristics. In the context of decision-making, fairness is the absence of any prejudice or favoritism toward an individual or group based on their inherent or acquired characteristics \cite{mehrabi2021survey}. Within the current technical framework, fairness is primarily defined as utilitarian fairness \cite{triantafyllopoulos2024enrolment} , which focuses on achieving equality in AI outputs across different groups through technical methods. While this definition is relatively narrow and does not fully address deeper issues such as structural inequalities or ethical fairness, it aligns with FAI's goal of comprehensive value alignment with humanity. A truly "friendly" AI system must embody fairness, treating all user groups equitably while maintaining harmony between technical and ethical considerations.

The technical application of fairness is currently focused on data, models and outputs:

\subsubsection{Data} 
Imbalanced data distributions often lead models to favour majority groups during training, thereby undermining fairness. Resampling techniques \cite{kim2020resampling} are commonly employed to adjust the proportions of minority and majority groups, resulting in a more balanced distribution. Additionally, generative approaches are utilised for data augmentation \cite{tran2021data}, increasing data diversity while mitigating bias during training. For example, generative models can produce additional samples for minority groups, improving their representation within the dataset.
\subsubsection{Model}

Model-level fairness applications primarily introduce constraints during training to adjust the learning process. For example, the Equalized Odds method \cite{tang2022attainability} adds constraints to the training objective, ensuring that model outputs are conditionally independent of protected attributes given the true labels. This constraint aims to prevent predictions from being biased by sensitive attributes across different groups. Additionally, recent generative methods remove features unrelated to the primary task, such as gender or age, and create new data representations to force the model to learn attribute-independent features, thereby reducing bias \cite{sun2024audio, tan2020improving}.

To address individual fairness, researchers propose incorporating constraints into training to ensure similar individuals receive similar predictions. For instance, "Fairness Through Awareness" \cite{dwork2012fairness} introduces distance-based similarity measures and incorporates these constraints during training to align predictions for similar individuals. Furthermore, "Accurate Fairness" \cite{li2023accurate} seeks to resolve the trade-off between individual fairness and model accuracy. This approach combines fairness and accuracy constraints during training, balancing these objectives and enhancing individual fairness without significant performance loss. These advancements represent important directions for improving fairness at the model level.

\subsubsection{Output}
At the output level, post-processing techniques adjust model predictions to meet fairness requirements without altering the model structure or requiring retraining. These methods modify outputs directly, achieving greater balance across different groups. Calibration techniques adjust prediction probabilities, such as setting different thresholds for various groups \cite{pfohl2022net}, to optimise result distribution and improve inter-group fairness. Distribution adjustment methods reallocate prediction labels or probabilities to reduce biases among groups. While these approaches offer high flexibility and applicability, they may lead to performance trade-offs and are less effective when addressing deep-rooted biases in data or the model itself.

To sum, fairness remains an important yet developing aspect of ANI. Current methods largely centre on utilitarian fairness, aiming to balance outcomes at group or individual levels. These methods are applied through pre-processing, in-processing, and post-processing techniques. In the context of FAI, fairness is both a fundamental requirement and a core component. It ensures alignment between AI systems and human values. Achieving comprehensive fairness enhances trust in AI and supports the broader goal of FAI: ethical and inclusive intelligence.
\subsection{Affective Computing}

Affective Computing (AC) is the study of systems designed to understand, interpret, and process human emotions \cite{picard1997affective}. It plays a crucial role in the development of FAI systems, as the ability to comprehend emotional cues is essential for creating interactions that are perceived as empathetic or "friendly." For instance, demonstrating kindness involves recognizing and responding to emotional signals, a domain where AC is particularly effective. In general, AC models emphasize emotion recognition, which is foundational for designing systems that can accurately perceive and respond to human affective states. The applications of AC span a wide range of fields, including healthcare, education, entertainment, and robotics, among others. In this section, we review the foundational developments and applications of AC, highlighting its potential as a key component in the design of FAI systems.

\subsubsection{Development history}
The development of AC has undergone significant historical progress. In its early stages, AC focused primarily on recognizing general emotional states, a field that shares overlap with psychology and cognitive science, for example with theories proposed by Ekman \cite{ekman1982felt}. As machine learning models advanced, AC evolved to emphasize emotion recognition, with key contributions such as the development of Facial Expression Recognition (FER) through the introduction of the Facial Action Coding System (FACS) by Ekman \cite{ekman1978facs} and its integration into emotion recognition technologies. Additionally, Speech Emotion Recognition (SER) emerged as a prominent application as well, utilizing acoustic features such as tone and pitch to predict emotional states proposed by Schuller \cite{schuller2003hidden}.

As the applications of AC have become increasingly practical, the field has evolved into specialized subfields, including the recognition of specific emotions and their application within particular domains. Healthcare provides a notable example of this evolution. AC has been integrated into physiological monitoring, utilizing biological sensors to assess emotional states by analyzing physiological features such as heart rate, skin responses, and brain activity \cite{hengameh2015biofeedback}. In their comprehensive review, Wang et al. \cite{wang2022systematicreviewaffectivecomputing} examine the breadth of affective computing models, while Singh et al. \cite{9707947} explore models tailored specifically to the psychological domain. Schuller et al. \cite{schuller2021review} also provide some future directions. Beyond healthcare, AC has also been applied to the analysis of everyday behaviors as indicators of emotional states \cite{wang2021affective}.

With the growing recognition of AC systems, their applications have expanded into multimodal frameworks. The shift from single-modality to multimodal approaches—such as integrating facial expressions, voice, and physiological signals like heart rate—has significantly advanced the field and was adept at analyzing complex emotions in intricate scenarios. Björn et al. \cite{schuller2002multimodal} present a multimodal approach to emotion recognition in audiovisual communication, providing valuable insights into the integration of visual and auditory cues for emotion detection in this domain. Shi and Huang \cite{shi2023multiemo} proposed a novel framework that effectively integrates multimodal cues by capturing cross-modal relationships to analyze various emotions in conversational contexts. Similarly, Jun et al. \cite{wu2024mlgat} developed a Multi-Layer Graph Attention Network, which aligns with similar applications in emotion recognition. Schuller et al. \cite{schuller2018multimodal} explored the analysis of user states and traits through multimodal data, advancing the paradigm in new directions. Moreover, recent reviews by Cabada et al. \cite{zatarain2023multimodal} highlight key challenges and opportunities, including the integration of LLMs into emotion recognition systems, marking a significant maturation in AC research. These developments suggest that AC models are approaching a level of sophistication where they can serve as essential components of FAI systems, enabling dynamic, real-time emotion analysis and understanding.

\subsubsection{Application}

After the vertical discussion of this fields development, we begin a horizontal review on AC application here. There we mainly review healthcare, education, entertainment, and robotics. 

\paragraph{Healthcare} Healthcare is one of the primary applications of AC, particularly in the domain of mental health. Liu et al. \cite{liu2024affective} present the most recent review of foundational models, challenges, and future directions in AC, building upon the valuable insights provided by Shu et al. \cite{shu2018review}. Neural networks are commonly employed in this domain \cite{dhuheir2021emotion}, and Mallol et al. \cite{mallol2020curriculum} introduce a curriculum-based approach as a notable example. AC models in healthcare typically focus on emotion recognition, integrating physiological signals measured by devices such as smartwatches. These models aim to monitor emotional states, enhance patient care, and support mental health management.

\paragraph{Education} The integration of AC is increasingly important in the educational process, particularly in virtual learning environments. Yang et al. \cite{yang2018emotion} propose models that detect emotions based on facial expressions, while Lasri et al. \cite{lasri2019facial} offer similar insights, emphasizing the use of convolutional neural networks. Additionally, sentiment analysis methods have proven valuable for emotion detection in educational contexts \cite{barron2019emotion}. Speech analysis models, as reviewed by Schröder \cite{schroder2001emotional}, are also beneficial for understanding emotional states in students. The field continues to evolve, as highlighted in the overview by Yu et al. \cite{yu2024bridging}. AC models in education are increasingly capable of monitoring remote teaching, particularly within the context of the current hybrid education model, where they can enhance both student engagement and learning outcomes.

\paragraph{Entertainment} AC models have significantly enhanced both commercial and non-commercial applications in the entertainment industry. De et al. \cite{de2023research} design an interface that connects AC models with brain-computer interactions, enabling more flexible and immersive experiences in entertainment contexts. Huang et al. \cite{huang2024decoding} utilize emotion decoding for movie classification, which can contribute to improving recommendation systems. Li \cite{li2025application} offers insights into the application of AC models in e-learning platforms, particularly using genetic algorithms for art design. AC models are capable of extracting nuanced emotional information in entertainment, supporting downstream tasks such as optimizing content for greater audience engagement and increasing media views.

\paragraph{Robotics} Robotics has made significant strides, but the integration of AC models can further enhance robots' empathy and reasoning capabilities. Spezialetti et al. \cite{spezialetti2020emotion} highlight recent advances and future perspectives in this area, while Stock et al. \cite{stock2022survey} provide comprehensive reviews, placing particular emphasis on the psychological aspects of robotics. Additional techniques have demonstrated the effectiveness of AC technologies in this domain. For instance, Leo et al. \cite{leo2015automatic} design a robot-child interaction system based on AC models, aimed at improving therapeutic outcomes, while Zhang et al. \cite{zhang2013intelligent} develop a humanoid robot that incorporates emotion recognition. Furthermore, some human-interactive robots, known as sociable agents, which are designed to demonstrate emotional behaviours and enhance human-robot interaction \cite{breazeal2003emotion}. These approaches enable FAI to be instructed with greater specificity, allowing it to analyse human emotions in a more targeted and precise manner. With the integration of AC models, robotics is positioned to become an increasingly promising field, empowering robots to better understand and respond to human emotions.

\subsubsection{Future direction}
From the early challenges in emotion detection, as developed by Schuller et al. \cite{schuller2013interspeech}, to the present, the field of AC remains a dynamic area of ongoing research. Schuller et al. \cite{schuller2024affective} emphasize that foundational models have reshaped the direction of AC research, suggesting that further integration of human psychological features is necessary. Additionally, the incorporation of empathy within AC models is becoming increasingly important, as some previous studies have explored \cite{ghandi2021embodied}, although significant gaps remain regarding how and what aspects of empathy should be embedded.

In summary, AC models hold considerable promise for fostering the development of FAI systems, much like their applications in the areas discussed above. These models are poised to become critical components of FAI applications, and the continued advancement of foundational AC technologies will play a key role in the evolution of these systems.

\section{Challenges and Suggestions}
\label{outlook}
In this section, we discuss the current challenges facing the FAI field and provide some forward-looking outlooks.
\subsection{Challenges}

One major obstacle lies in the inability to establish a unified definition for FAI. Theoretical definitions of `friendliness' vary widely, with scholars proposing conflicting frameworks. For instance, Yudkowsky’s CEV supports humanity’s ideal collective will, while Mittelstadt’s ``not harm humanity'' principle prioritises safety and harm avoidance. These differing perspectives lead to disagreements over the scope and objectives of FAI, further complicating efforts to quantify its success. Without a universally accepted definition, practitioners lack clear guidelines for evaluating whether an AI system aligns with FAI principles, rendering its practical application even more challenging.
The absence of a widely recognised definition for FAI has led to fragmented discourse on the topic. While many scholars share similar theoretical concerns about the future of AI and have proposed related ideas, they often do not explicitly use the term `FAI' \cite{livingston2019future}. This lack of consistent terminology limits the dissemination of these ideas, making it challenging to collect and organise relevant contributions into a coherent body of knowledge. As a result, the field struggles to achieve the critical mass needed to drive large-scale collaborative efforts.

Also, this ambiguity stems from defining `friendliness' in a culturally diverse world. Different societies and cultures prioritise distinct moral values; for example, individual autonomy may be emphasised in some cultures, while collective welfare is prioritised in others. This raises the question of how to balance these conflicting priorities within a single framework. Additionally, determining what values should take precedence in cross-cultural contexts is an inherently subjective process, further complicating efforts to create universally friendly AI systems. To meet these demands, FAI must not only accommodate diverse ethical standards but also navigate complex decisions about whose interests to prioritise, a task that remains unresolved in theory and practice.
As previously discussed, FAI lacks a unified theoretical framework. Consequently, developers cannot rely on clear guidelines when designing and implementing AI systems, resulting in confusion and a lack of direction in practice. Developers find it challenging to determine whether a specific technological implementation aligns with the core goals of FAI. In practical scenarios, this ambiguity in ethical standards may lead to further difficulties. Moreover, the absence of a clear theoretical foundation makes it nearly impossible to quantify or evaluate the success of ideal FAI systems.

Furthermore, The development of FAI remains largely a preparatory effort, as the AGI era has not yet arrived. Most current research focuses on ANI technologies. While progress has been made in areas like computer vision and natural language processing, some fields still fall short of human-level capabilities. For example, speech generation often lacks nuance and contextual understanding. Tasks requiring fine motor skills or complex reasoning, such as robotic surgery or decision-making in uncertain environments, also often remain beyond AI’s abilities. As a result, research today focuses on building more precise and efficient AI models rather than addressing broader ethical alignment or value-driven frameworks. This gap highlights the difficulty of transitioning from task-specific ANI to the comprehensive systems required for FAI.

Besides, it is unclear whether some current AI subfields will eventually be formally included within the FAI framework. For instance, while XAI aims to enhance transparency and trustworthiness, its connection to FAI’s broader ethical goals is mostly indirect. Privacy-preserving technologies, although effective in building trust, primarily address technical concerns and lack full integration with FAI’s long-term objectives. Similarly, AC seeks to improve AI’s ability to understand and respond to human emotions, but its focus remains on human-machine interaction rather than deeper alignment with human values. Consequently, no current research provides a systematic definition of the technical directions that should or could be included under FAI. This ambiguity not only affects the prioritisation of research efforts but also leads to a fragmented approach to long-term goals.

The realisation of FAI requires broad collaboration across disciplines and industries, yet this collaboration faces multiple challenges in practice. First, differences in regulations and policy priorities between countries create significant obstacles. Some nations prioritise technological competitiveness, while others focus on ethics and safety. This lack of alignment complicates the creation of global standards. Second, the sensitive nature of AI technology adds another layer of complexity. In areas such as national security, military applications, or biomedical research, the sharing of data and technologies is heavily restricted. This often leads to a lack of trust among stakeholders. Third, businesses and academic institutions often have diverging objectives in collaboration. Companies are generally driven by commercialisation and rapid application, whereas academics aim to explore long-term ethical and theoretical questions. These differing priorities limit resource allocation and reduce the efficiency of joint efforts. Finally, the globalisation of AI development exacerbates challenges, as cultural, economic, and political differences further complicate cooperation. For instance, some regions emphasise stricter privacy protections, while others prioritise freedom in technological innovation. Together, these factors make the multi-stakeholder collaboration required for FAI highly complex when the AGI era lands.
\subsection{Suggestions}
Building on these challenges, we propose several potential suggestions to address the obstacles of FAI.

\subsubsection{Establishing a Unified Definition Framework}
The lack of a unified definition for FAI remains a significant obstacle to both theoretical exploration and practical application. As discussed in Section \ref{debate}, much of the criticism directed towards FAI is not against its core ideas (e.\,g., value alignment and deontology) but stems from concerns over its ambiguous scope and perceived impracticality due to its evolving nature. To address this, we propose that international academic institutions lead efforts to convene ethicists, policymakers, and AI researchers to develop a modular framework. This framework should clearly define the core principles and boundaries of FAI. Additionally, instead of dismissing FAI as unachievable, it is important to address these concerns through incremental progress. By first agreeing on a theoretical foundation and then considering practical implementation strategies, the ideal of FAI can be approached systematically rather than prematurely dismissed.

\subsubsection{Consolidating Fragmented Knowledge Systems}
The lack of consistency in terminology and research directions has led to the fragmentation of FAI-related knowledge, limiting its ability to achieve large-scale impact. To address this, we propose the creation of an open-access knowledge-sharing platform. Such a platform, akin to an `FAI Wiki' could centralise academic papers, technological advancements, and industry practices in one accessible location. By incorporating features such as multilingual translation and systematic categorisation, this platform would facilitate collaboration among researchers and foster knowledge accumulation. It would provide a solid foundation for advancing FAI research and ensuring its broader development.

\subsubsection{Developing a Cross-Cultural Ethical Framework}
The concept of `friendliness' varies across cultures, reflecting diverse moral values and social priorities. To address this complexity, we propose a multi-layered ethical decision-making system that combines global core principles, such as fairness and privacy, with dynamic adaptability to regional ethical requirements. For instance, an AI system could prioritise individual autonomy in societies where personal freedom is emphasised while focusing on collective welfare in cultures that value community-oriented decision-making. By tailoring ethical decisions to the specific moral priorities of each region, while maintaining consistency with overarching standards, such a framework allows AI systems to align more effectively with human values.

\subsubsection{Accelerating Breakthroughs in Technical Capabilities}
Many ANI systems face limitations in areas such as nuanced expression, long-memory storage, and adaptive decision-making. These gaps highlight the need for targeted advancements to bridge the transition from ANI to AGI and ensure alignment with FAI principles. To address these challenges, research efforts should focus on enhancing AI’s ability to process and utilise contextual information over extended periods, enabling systems to handle complex and dynamic environments more effectively. For example, enhancing memory architectures can improve long-term context retention, while refining natural language models can enable AI to better understand and respond to cultural and emotional nuances, fostering more meaningful interactions.

However, while pursuing these innovations, it is essential to balance technical advancements with continuous evaluation of their broader implications \cite{shen2023large}. Focusing solely on performance risks overlooking other aspects such as ethical considerations and alignment with human values. Developers should integrate regular assessments of how new technologies impact societal norms and ethical frameworks, ensuring that progress in capabilities does not come at the expense of responsible and aligned AI development.

\subsubsection{Defining potential subfields for FAI}

To address the ambiguity over which AI subfields belong within the FAI framework, we propose identifying and categorising technologies relevant to FAI. In our discussion above, XAI, privacy, fairness and AC are promising candidates, as they provide foundational tools for transparency, trust, and responsiveness to human needs. While these fields alone cannot fully encompass the scope of FAI, they offer significant contributions to its framework. Additionally, other subfields that have not been explicitly discussed here may also play a important role and should be considered in future explorations. We recommend that academic bodies or professional associations take the lead in defining and formalising the scope of FAI-related technologies. By systematically identifying and including such technologies, the field can establish a clearer trajectory for integrating ethical principles into technological development and ensure a more unified approach toward FAI.

\subsubsection{Promoting Multi-Stakeholder Collaboration}
To address the complexities of multi-stakeholder collaboration in FAI conception,  establishing structured mechanisms is important to reconcile diverse priorities across disciplines, industries, and nations. An international coordination body, potentially led by organisations such as the United Nations, could provide a platform to harmonise differing interests, including governmental emphasis on safety, corporate focus on commercialisation, and academic dedication to ethical and theoretical considerations. By developing clear cooperation guidelines, standardising resource-sharing frameworks, and fostering mutual trust through transparent practices, such an initiative could effectively bridge these divides. 

\subsubsection{Enhancing Public Trust and Awareness}  
The ultimate goal of FAI is to ensure mutual trust and respect between humans and AI. Gaining public support and confidence in AI systems is essential for achieving this aim. Promoting AI requires transparent communication about its objectives and benefits. Educational initiatives, such as interactive exhibits or accessible online courses, can help the public understand and appreciate the technology. Additionally, involving the public in discussions about FAI’s ethical principles and potential applications can create a sense of inclusion, fostering greater trust and collective support for its development.

\section{Conclusion}

In this paper, we have conducted a comprehensive review of FAI, primarily from an ethical perspective. We have summarised public perspectives both supporting and opposing the development of FAI, alongside a discussion of its formal definition, which we have presented in a clear and accessible format. Additionally, we explored relevant FAI applications, focusing on XAI, privacy, and AC. Furthermore, we outlined the challenges associated with current technological advancements and discussed potential future directions for the field. In conclusion, we advocate for the continued development of FAI and emphasise its value, expressing our strong support for its advancement.

\section*{Acknowledgment}

\bibliographystyle{IEEEtran}
\bibliography{refs}

\end{document}